\documentclass[11pt,a4paper]{article}
\usepackage[hyperref]{emnlp2018}
\usepackage{times}
\usepackage{latexsym}
\usepackage{todonotes}
\usepackage{basecommon}
\usepackage{url}
\usepackage{amsmath}
\usepackage{booktabs}
\usepackage[font={small}]{caption}
\usepackage{multirow}
\usepackage{xcolor}

\aclfinalcopy 
\def\aclpaperid{***} 

\title{Learning Neural Templates for Text Generation}

 \author{Sam Wiseman \quad \quad \quad  Stuart M. Shieber  \quad \quad \quad Alexander M. Rush \\
         School of Engineering and Applied Sciences \\ Harvard University \\ Cambridge, MA, USA \\ {\tt \{swiseman,shieber,srush\}@seas.harvard.edu }}

\date{}

\begin{document}
\maketitle
\begin{abstract}
 While neural, encoder-decoder models have had significant empirical success in text generation, there remain several unaddressed problems with this style of generation. Encoder-decoder models are largely (a) uninterpretable, and (b) difficult to control in terms of their phrasing or content. This work proposes a neural generation system using a hidden semi-markov model (HSMM) decoder, which learns latent, discrete templates jointly with learning to generate. We show that this model learns useful templates, and that these templates make generation both more interpretable and controllable. Furthermore, we show that this approach scales to real data sets and achieves strong performance nearing that of encoder-decoder text generation models.
\end{abstract}

\section{Introduction}

With the continued success of encoder-decoder models for machine translation and related tasks, there has been great interest in extending these methods to build general-purpose, data-driven natural language generation (NLG) systems~\citep{mei2016what,duvsek2016sequence,lebret2016neural,chisholm2017learning,wiseman2017challenges}. These encoder-decoder models~\citep{sutskever2014sequence,cho2014on,bahdanau2015neural} use a neural encoder model to represent a source knowledge base, and a decoder model to emit a textual description word-by-word, conditioned on the source encoding. This style of generation contrasts with the more traditional division of labor in NLG, which famously emphasizes addressing the two questions of ``what to say'' and ``how to say it'' separately, and which leads to systems with explicit content selection, macro- and micro-planning, and surface realization components~\citep{reiter1997building,jurafsky2014speech}.

\begin{figure}[h]
\centering
\fbox{
\parbox{0.96\linewidth}{
\textbf{Source Entity}: Cotto
\vspace{0.1cm}

type[coffee shop], rating[3 out of 5],\\
food[English], area[city centre], \\
price[moderate],  near[The Portland Arms]

\vspace{0.3cm}

\textbf{System Generation:}

\underline{Cotto}  is a  \underline{coffee shop}  serving  \underline{English} food in the  \underline{moderate} price range.  It is located near \underline{The Portland Arms}. Its customer rating is  \underline{3 out of 5}.

\vspace{0.3cm}

\textbf{Neural Template:}
    \vspace{0.3cm}
    
$\textcolor{gray}{|} $ $ \substack{\text{The \rule{15pt}{1pt}}\\\text{\rule{15pt}{1pt}}\\ \dots}
\ \textcolor{gray}{|} \  $ $ \substack{\text{is a}\\\text{is an}\\\text{is an expensive}\\ \dots}
\ \textcolor{gray}{|} $ $ \text{\rule{15pt}{1pt}} \ \textcolor{gray}{|} \  $ $ \substack{\text{providing}\\\text{serving}\\\text{offering}\\ \dots}
\ \textcolor{gray}{|} $ $ \text{\rule{15pt}{1pt}} $ \\ $\textcolor{gray}{|} \  $ $ \substack{\text{food}\\\text{cuisine}\\\text{foods}\\ \dots}
\ \textcolor{gray}{|} \  $ $ \substack{\text{in the}\\\text{with a}\\\text{and has a}\\ \dots} \ \textcolor{gray}{|} $ $ \text{\rule{15pt}{1pt}} \
\ \textcolor{gray}{|} \  $ $ \substack{\text{price range}\\\text{price bracket}\\\text{pricing}\\ \dots}
\ \textcolor{gray}{|} \  $ $ \text{.}$ 
$\textcolor{gray}{|}$ $ \substack{\text{It's}\\\text{It is}\\\text{The place is}\\ \dots} \ $ 

$\textcolor{gray}{|} \  $ $ \substack{\text{located in the}\\\text{located near}\\\text{near}\\ \dots}
\ \textcolor{gray}{|} $ $ \text{\rule{15pt}{1pt}} \ \textcolor{gray}{|} \  $ $ \text{.} $ $ \textcolor{gray}{|} \  $ $ \substack{\text{Its customer rating is}\\\text{Their customer rating is}\\\text{Customers have rated it}\\ \dots}
\ \textcolor{gray}{|} $ $ \text{\rule{15pt}{1pt}} \ \textcolor{gray}{|} $ . 
}}
    \caption{An example template-like generation from the E2E Generation dataset~\citep{novikova2017e2e}. Knowledge base $x$ (top) contains 6 records, and $\hat{y}$ (middle) is a system generation; records are shown as \texttt{type[value]}.
    An induced neural template (bottom) is learned by the system and employed in generating $\hat{y}$. Each cell represents a segment in the learned segmentation, and ``blanks'' show where slots are filled through copy attention during generation.
    }
    \label{fig:e2eex}
\end{figure}

Encoder-decoder generation systems appear to have increased the fluency of NLG outputs, while reducing the manual effort required. However, due to the black-box nature of generic encoder-decoder models, these systems have also largely sacrificed two important desiderata that are often found in more traditional systems, namely (a) interpretable outputs that (b) can be easily controlled in terms of form and content. 


This work considers building interpretable and controllable neural generation systems, and proposes a specific first step: a new data-driven generation model for learning discrete, \textit{template-like} structures for conditional text generation. The core system uses a novel, neural hidden semi-markov model (HSMM) decoder, which provides a principled approach to template-like text generation. We further describe efficient methods for training this model in an entirely data-driven way by backpropagation through inference. 
Generating with the template-like structures induced by the neural HSMM allows for the explicit representation of what the system intends to say (in the form of a learned template) and how it is attempting to say it (in the form of an instantiated template). 

We show that we can achieve performance competitive with other neural NLG approaches, while making progress satisfying the above two desiderata. Concretely, our experiments indicate that we can induce explicit templates (as shown in Figure 1) while achieving competitive automatic scores, and that we can control and interpret our generations by manipulating these templates. Finally, while our experiments focus on the data-to-text regime, we believe the proposed methodology represents a compelling approach to learning discrete, latent-variable representations of conditional text.

\section{Related Work}
\label{sec:relatedwork}
A core task of NLG is to generate textual descriptions of knowledge base records. A common approach is to use hand-engineered templates~\citep{kukich1983design,mckeown1992text,mcroy2000yag}, but there has also been interest in creating templates in an automated manner. For instance, many authors induce templates by clustering sentences and then abstracting templated fields with hand-engineered rules~\citep{angeli2010simple,kondadadi2013statistical,howald2013domain}, or with a pipeline of other automatic approaches~\citep{wang2013domain}. 

There has also been work in incorporating probabilistic notions of templates into generation models~\citep{liang2009learning,konstas2013global}, which is similar to our approach. However, these approaches have always been conjoined with discriminative classifiers or rerankers in order to actually accomplish the generation~\citep{angeli2010simple,konstas2013global}. In addition, these models explicitly model knowledge base field selection, whereas the model we present is fundamentally an end-to-end model over generation segments.

Recently, a new paradigm has emerged around neural text generation systems based on machine translation~\citep{sutskever2014sequence,cho2014on,bahdanau2015neural}. Most of this  work has used unconstrained black-box encoder-decoder approaches. There has been some work on discrete variables in this context, including extracting representations~\citep{shen2018neural}, incorporating discrete latent variables in text modeling~\citep{yang2018breaking}, and using non-HSMM segmental models for machine translation or summarization~\citep{yu2016online,wang2017sequence, huang2018towards}. \citet{dai2017recurrent} develop an approximate inference scheme for a neural HSMM using RNNs for continuous emissions; in contrast we maximize the exact log-marginal, and use RNNs to parameterize a discrete emission distribution. Finally, there has also been much recent interest in segmental RNN models for non-generative tasks in NLP~\citep{tang2016end,kong2016segmental,lu2016segmental}. 

The neural text generation community has also recently been interested in ``controllable'' text generation~\citep{hu2017controllable}, where various aspects of the text (often sentiment) are manipulated or transferred~\citep{shen2017style,zhao2017adversarially,li2018style}. 
In contrast, here we focus on controlling either the content of a generation or the way it is expressed by manipulating the (latent) template used in realizing the generation.

\section{Overview: Data-Driven NLG}
\begin{figure}[t]
\centering
\begin{tabular}{@{}l@{}}
\toprule
\begin{tabular}[c]{c}
\hspace{0.6cm}
\includegraphics[scale=1.2]{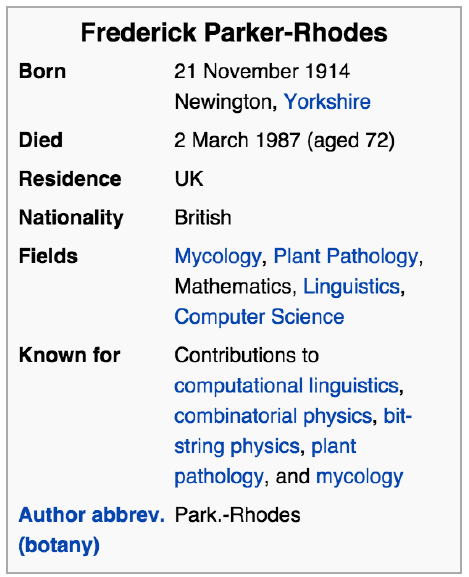} 
\end{tabular} \\
\midrule
\begin{tabular}[c]{@{}l@{}}
{\small Frederick Parker-Rhodes (21 March 1914 - 21 November} \\ 
{\small 1987) was an English linguist, plant pathologist, computer} \\ 
{\small scientist, mathematician, mystic, and mycologist.} 
\end{tabular} \\
\bottomrule
\end{tabular}
\caption{An example from the WikiBio dataset~\citep{lebret2016neural}, with a database $x$ (top) for Frederick Parker-Rhodes and corresponding reference generation $y$ (bottom).}
\label{fig:wbex}
\end{figure}

Our focus is on generating a textual description of a knowledge base or 
meaning representation. Following standard notation \citep{liang2009learning,wiseman2017challenges}, let $x \niceq \{r_1 \ldots r_J\}$ be a collection of records. A record is made up of a \textit{type} ($r.t$), an \textit{entity} ($r.e$), 
and a \textit{value} ($r.m$).  For example, a knowledge base of restaurants might have a record with $r.t$ = \texttt{Cuisine}, $r.e$ = \texttt{Denny's}, and $r.m$ = \texttt{American}. The aim is to generate an adequate and fluent text description $\hat{y}_{1:T} \niceq \hat{y}_1, \ldots, \hat{y}_T$ of $x$. Concretely, we consider the E2E Dataset~\citep{novikova2017e2e} and the WikiBio Dataset~\citep{lebret2016neural}. We show an example E2E knowledge base $x$ in the top of Figure~\ref{fig:e2eex}. The top of Figure~\ref{fig:wbex} shows an example knowledge base $x$ from the WikiBio dataset, where it is paired with a \textit{reference} text $y \niceq y_{1:T}$ at the bottom.

The dominant approach in neural NLG has been to use an encoder network over $x$ and then a conditional decoder network to generate $y$, training the whole system in an end-to-end manner. To generate a description for a given example, a black-box network (such as an RNN) is used to produce a distribution over the next word, from
which a choice is made and fed back into the system. The entire distribution is driven by the internal states of the neural network. 

While effective, relying on a neural decoder makes it difficult to understand what aspects of $x$ are correlated with a particular system output. This leads to problems both in controlling fine-grained aspects of the generation process and in interpreting model mistakes. 

As an example of why controllability is important, consider the records in Figure~\ref{fig:e2eex}. Given these inputs  an end-user might want to generate an output meeting specific constraints, such as not mentioning any information relating to customer rating. Under a standard encoder-decoder style model, one could filter out this information 
either from the encoder or decoder, but in practice this would lead to unexpected changes in output that 
might propagate through the whole system.


As an example of the difficulty of interpreting mistakes, consider the following actual generation from an encoder-decoder style system for the records in Figure~\ref{fig:wbex}: "frederick parker-rhodes (21 november 1914 - 2 march 1987) was an english mycology and plant pathology, mathematics at the university of uk."
In addition to not being fluent, it is unclear what the end of this sentence is even attempting to convey: it  may be attempting to convey a fact not actually in the knowledge base (e.g., where Parker-Rhodes studied), or perhaps it is simply failing to fluently realize information that \textit{is} in the knowledge base (e.g., Parker-Rhodes's country of residence). 

Traditional NLG systems~\citep{kukich1983design,mckeown1992text,belz2008automatic,gatt2009simplenlg}, in contrast, largely avoid these problems. Since they typically employ an explicit planning component, which decides which knowledge base records to focus on, and a surface realization component, which realizes the chosen records, the intent of the system is always explicit, and it may be modified to meet constraints.


The goal of this work is to propose an approach to neural NLG that addresses these issues in a principled way. 
We target this goal by proposing a new model that generates with template-like objects induced by a neural HSMM (see Figure~\ref{fig:e2eex}). Templates are useful here because they represent a fixed plan for the generation's content, and because they make it clear what part of the generation is associated with which record in the knowledge base.

\section{Background: Semi-Markov Models}
What does it mean to learn a template? It is natural to think of a template as a sequence of typed text-segments, perhaps with some segments acting as the template's ``backbone''~\citep{wang2013domain}, and the remaining segments filled in from the knowledge base. 

A natural probabilistic model conforming with this intuition is the hidden semi-markov model (HSMM)~\citep{gales1993theory,ostendorf1996hmm}, which models latent segmentations in an output sequence. 
Informally, an HSMM is much like an HMM, except emissions may last multiple time-steps, and multi-step emissions need not be independent of each other conditioned on the state. 

We briefly review HSMMs following \citet{murphy2002hidden}. Assume we have a sequence of observed tokens $y_{1} \ldots y_{T}$ and a discrete, latent state $z_t \nicein \{1,\ldots,K\}$ for each timestep. We additionally use two per-timestep variables to model multi-step segments: a length variable $l_t \nicein \{1,\ldots, L\}$ specifying the length of the current segment, and a deterministic binary variable $f_t$ indicating whether a segment finishes at time $t$. We will consider in particular \textit{conditional} HSMMs, which condition on a source $x$, essentially giving us an HSMM decoder.





An HSMM specifies a joint distribution on the observations and latent segmentations. Letting $\theta$ denote all the parameters of the model, and using the variables introduced above, we can write the corresponding joint-likelihood as follows
 \begin{align*}
p(y, z, l, f \given x; \theta) = &\prod_{t=0}^{T-1} p(z_{t+1}, l_{t+1} \given z_{t}, l_{t}, x)^{f_t} \\
\times &\prod_{t=1}^T p(y_{t-l_t+1:t} \given z_t, l_t, x)^{f_t},
\end{align*}
where we take $z_0$ to be a distinguished start-state, and the deterministic $f_t$ variables are used for excluding non-segment log probabilities.
We further assume $p(z_{t+1}, l_{t+1} \given z_{t}, l_{t}, x)$ factors as $p(z_{t+1} \given z_{t}, x) \times p(l_{t+1} \given z_{t+1})$. Thus, the likelihood is given by the product of the probabilities of each discrete state transition made, the probability of the length of each segment given its discrete state, and the probability of the observations in each segment, given its state and length.



\section{A Neural HSMM Decoder}
\label{sec:model}
We use a novel, neural parameterization of an HSMM
to specify the probabilities in the likelihood above. 
This full model, sketched out in Figure~3, allows us to incorporate
the modeling components, such as LSTMs and attention, that make neural text generation
effective, while maintaining the HSMM structure.

\subsection{Parameterization}

Since our model must condition on $x$, let $\boldr_j \nicein \reals^d$ represent a real embedding of record $r_j \nicein x$, and let $\boldx_a \nicein \reals^d$ represent a real embedding of the entire knowledge base $x$, obtained by max-pooling coordinate-wise over all the $\boldr_j$. It is also useful to have a representation of just the unique \textit{types} of records that appear in $x$, and so we also define $\boldx_u \nicein \reals^d$ to be the sum of the embeddings of the unique types appearing in $x$, plus a bias vector and followed by a ReLU nonlinearity.


\vspace{-0.15cm}
 \paragraph{Transition Distribution} The transition distribution $p(z_{t+1} \given z_{t}, x)$ may be viewed as a $K \, {\times} \, K$ matrix of probabilities, where each row sums to 1. We define this matrix to be 
\begin{align*}
    p(z_{t+1} \given z_{t}, x) \propto \boldA \boldB + \boldC(\boldx_u) \boldD(\boldx_u),
\end{align*}
where $\boldA \nicein \reals^{K \times m_1}$, $\boldB \nicein \reals^{m_1 \times K}$ are state embeddings, and where $\boldC : \reals^d \rightarrow \reals^{K \times m_2}$ and $\boldD : \reals^d \rightarrow \reals^{m_2 \times K}$ are parameterized non-linear functions of $\boldx_u$. We apply a row-wise $\mathrm{softmax}$ to the resulting matrix to obtain the desired probabilities. 

\vspace{-0.15cm}
\paragraph{Length Distribution}
We simply fix all length probabilities $p(l_{t+1} \given z_{t+1})$ to be uniform up to a maximum length $L$.\footnote{We experimented with parameterizing the length distribution, but found that it led to inferior performance. Forcing the length probabilities to be uniform encourages the model to cluster together functionally similar emissions of different lengths, while parameterizing them can lead to states that specialize to specific emission lengths.}  

\begin{figure}
    \centering
    \begin{tikzpicture}[node distance=0.6cm]
    \draw [step=0.2cm,gray,very thin] (-1.11, -1.11) grid  (1.1, 1.11);
    \node [draw, circle, fill=black!10](x){$x$} ;
    \node[circle, draw, above right=of x, xshift = 1cm, fill=red!20](z){$z_1$};
    \node[ draw, rounded corners, below=of z, fill=red!10](rnn){RNN};
    \node[circle, draw, below= of rnn](y){$y_1$};
    \node[circle, draw, right= of y](ya){$y_2$};
    \node[circle, draw, right= of ya](yb){$y_3$};
    \node[circle, draw, right= of yb](yc){$y_4$};
    \node[draw, rounded corners,  above= of yc, fill=blue!10](rnnb){RNN};
    \node[circle, draw, above= of rnnb, fill=blue!20](zb){$z_{4}$};
    \draw[-] (z) -- (rnn) -- (y);
    \draw[-]  (rnn) -- (ya);
    \draw[-] (rnn) -- (yb);

    \draw[-] (zb) -- (rnnb) -- (yc);
     \draw (z) --node(mlp)[fill=white, draw, rounded corners]{T} (zb);
     \draw (x.north) edge [bend left=40] (mlp);
     \draw (x) edge[] (rnn);
     \draw (x) edge[bend left=20] (rnnb);
     
    \end{tikzpicture}

    \caption{HSMM factor graph (under a known segmentation) to illustrate parameters. Here we assume $z_1$ is in the ``red'' state (out of $K$ possibilities), and transitions to the ``blue'' state after emitting three words. The transition model, shown as $T$, is a function of the two states and the neural encoded source $x$. The emission model is a function of a ``red'' RNN model (with copy attention over $x$) that generates words 1, 2 and 3. After transitioning, the next word $y_4$ is generated by the ``blue'' RNN, but independently of the previous words.}
    \label{fig:my_label}
\end{figure}
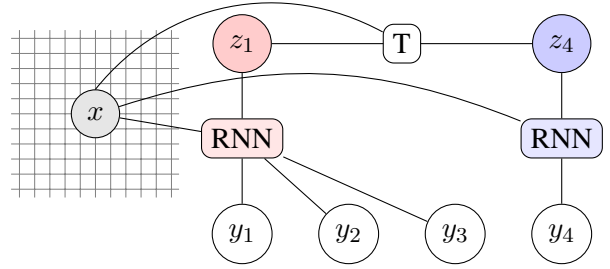

\vspace{-0.15cm}
\paragraph{Emission Distribution}
The emission model models the generation of a text segment conditioned on 
a latent state and source information, and so requires a richer parameterization. 
Inspired by the models used for neural NLG, we base this model on an RNN decoder, and write a segment's probability as a product over token-level probabilities,
\begin{align*}
    p&(y_{t-l_t+1:t} \given z_t \niceq k , l_t \niceq l, x) = \\
    &\prod_{i=1}^{l_t} p(y_{t-l_t+i} \given y_{t-l_t+1:t-l_t+i-1}, z_t \niceq k, x) \\
    &\times p({<}\text{/seg}{>} \given y_{t-l_t+1:t}, z_t \niceq k, x) \times \mathbf{1}_{\{l_t \niceq l\}},
\end{align*}
where ${<}\text{/seg}{>}$ is an end of segment token. The RNN decoder uses attention and copy-attention over the embedded records $\boldr_j$, and is conditioned on $z_t \niceq k$ by concatenating an embedding corresponding to the $k$'th latent state to the RNN's input; the RNN is also conditioned on the entire $x$ by initializing its hidden state with $\boldx_a$.




More concretely, let $\boldh^k_{i-1} \nicein \reals^d$ be the state of an RNN conditioned on $x$ and $z_t \niceq k$ (as above) run over the sequence $y_{t-l_t+1:t-l_t+i-1}$. 
We let the model attend over records $\boldr_j$ using $\boldh^k_{i-1}$ (in the style of \citet{luong2015effective}), 
producing a context vector $\boldc^k_{i-1}$.  We may then obtain scores $\boldv_{i-1}$ for each word in the output vocabulary, 
\begin{align*}
\boldv_{i-1} \niceq \boldW \tanh(\boldg_1^k \circ [\boldh^k_{i-1}, \boldc^k_{i-1}]) ,
\end{align*}
with parameters $\boldg_1^k \nicein \reals^{2d}$ and $\boldW \nicein \reals^{V \times 2d}$. Note that there is a $\boldg_1^k$ vector for each of $K$ discrete states. To additionally implement a kind of slot filling, we allow  emissions to be directly copied from the value portion of the records $r_j$ using copy attention~\citep{gulcehre2016pointing,gu2016incorporating,yang2016reference}. Define copy scores,
\begin{align*}
    \rho_j = \boldr_j^{\trans} \tanh(\boldg_2^k \circ \boldh^k_{i-1}),
\end{align*}
where $\boldg_2^k \nicein \reals^{d}$. 
We then normalize the output-vocabulary and copy scores together, to arrive at
\begin{align*}
    \widetilde{\boldv}_{i-1} \niceq \mathrm{softmax}([\boldv_{i-1}, \rho_1, \ldots, \rho_J]),
\end{align*}
and thus
\begin{align*}
    p(y_{t-l_t+i} &\niceq w \given y_{t-l_t+1:t-l_t+i-1}, z_t \niceq k, x) = \\
    &\widetilde{\boldv}_{i-1, w} + \sum_{j: r_j.m \niceq w} \widetilde{\boldv}_{i-1, V+j}.
\end{align*}

\vspace{-0.15cm}
\paragraph{An Autoregressive Variant} The model as specified assumes segments are independent conditioned on the associated latent state and $x$. While this assumption still allows for reasonable performance, we can tractably allow interdependence between tokens (but not segments) by having each next-token distribution depend on all the previously generated tokens, giving us an autoregressive HSMM. For this model, we will in fact use $p(y_{t-l_t+i} \niceq w \given y_{1:t-l_t+i-1}, z_t \niceq k, x)$ in defining our emission model, which is easily implemented by using an additional RNN run over all the preceding tokens. We will report scores for both non-autoregressive and autoregressive HSMM decoders below.

\subsection{Learning}
\label{sec:learning}
The model requires fitting a large set of neural network parameters. 
Since we assume $z$, $l$, and $f$ are unobserved, we marginalize over these variables to maximize the log marginal-likelihood of the observed tokens $y$ given $x$. 
The HSMM marginal-likelihood calculation can be carried out efficiently with a dynamic program analogous to either the forward- or backward-algorithm familiar from HMMs~\citep{rabiner1989tutorial}. 

It is actually more convenient to use the backward-algorithm formulation when using RNNs to parameterize the emission distributions, and we briefly review the backward recurrences here, again following~\citet{murphy2002hidden}. We have:
\begin{align*}
\beta_t(j) &= p(y_{t+1:T} \given z_t \niceq j, f_t \niceq 1, x)\\
&= \sum_{k=1}^K \beta^{*}_t(k) \, p(z_{t+1} \niceq k \given z_t = j) \\
\beta^{*}_t(k) &= p(y_{t+1:T} \given z_{t+1}=k, f_t=1, x)\\
&= \sum_{l=1}^L \Big[ \beta_{t+l}(k) \, p(l_{t+1} \niceq l \given z_{t+1} \niceq k) \\ 
& \quad \quad \quad \; p(y_{t+1:t+l} \given z_{t+1} \niceq k, l_{t+1} \niceq l) \Big], 
\end{align*}
with base case $\beta_T(j) \niceq 1$. We can now obtain the marginal probability of $y$ as $p(y \given x) \niceq \sum_{k=1}^K \beta^{*}_0(k) \, p(z_1 \niceq k)$,
where we have used the fact that $f_0$ must be 1, and we therefore train to maximize the log-marginal likelihood of the observed $y$:
\begin{align} \label{eq:marginal}
    \ln p(y \given x; \theta) = \ln \sum_{k=1}^K \beta^{*}_0(k) \, p(z_1 \niceq k).
\end{align}

Since the quantities in~\eqref{eq:marginal} are obtained from a dynamic program, which is itself differentiable, we may simply maximize with respect to the parameters $\theta$ by back-propagating through the dynamic program; this is easily accomplished with automatic differentiation packages, and we use \texttt{pytorch}~\citep{paszke2017automatic} in all experiments.

\subsection{Extracting Templates and Generating}
\label{sec:segmentation}
After training, we could simply condition on a new database and generate with beam search, as is standard with encoder-decoder models. However, the structured approach we have developed allows us to generate in a more template-like way, giving us more interpretable and controllable generations. 

First, note that given a database $x$ and reference generation $y$ we can obtain the MAP assignment to the variables $z$, $l$, and $f$ with a dynamic program similar to the Viterbi algorithm familiar from HMMs. These assignments will give us a typed segmentation of $y$, and we show an example Viterbi segmentation of some training text in Figure~\ref{fig:segmentation}. Computing MAP segmentations allows us to associate text-segments (i.e., phrases) with the discrete labels $z_t$ that frequently generate them. These MAP segmentations can be used in an exploratory way, as a sort of dimensionality reduction of the generations in the corpus. More importantly for us, however, they can also be used to guide generation.

\begin{figure}[t!]
    [The Golden Palace]$_{55}$ [is a]$_{59}$ [coffee shop]$_{12}$ [providing]$_{3}$ [Indian]$_{50}$ [food]$_{1}$ [in the]$_{17}$ [\pounds 20-25]$_{26}$ [price range]$_{16}$ [.]$_{2}$ [It is]$_{8}$ [located in the]$_{25}$ [riverside]$_{40}$ [.]$_{53}$ [Its customer rating is]$_{19}$ [high]$_{23}$ [.]$_{2}$
    \caption{A sample Viterbi segmentation of a training text; subscripted numbers indicate the corresponding latent state. From this we can extract a template with $S \niceq 17$ segments; compare with the template used at the bottom  of Figure~\ref{fig:e2eex}. }
    \label{fig:segmentation}
\end{figure}

In particular, since each MAP segmentation implies a sequence of hidden states $z$, we may run a \textit{template extraction} step, where we collect the most common ``templates'' (i.e., sequences of hidden states) seen in the training data. Each ``template'' $z^{(i)}$ consists of a sequence of latent states, with $z^{(i)} \niceq z^{(i)}_1, \ldots z^{(i)}_S$ representing the $S$ distinct segments in the $i$'th extracted template (recall that we will technically have a $z_t$ for each time-step, and so $z^{(i)}$ is obtained by collapsing adjacent $z_t$'s with the same value); see Figure~\ref{fig:segmentation} for an example template (with $S \niceq 17$) that can be extracted from the E2E corpus. The bottom of Figure~\ref{fig:e2eex} shows a visualization of this extracted template, where discrete states are replaced by the phrases they frequently generate in the training data.

With our templates $z^{(i)}$ in hand, we can then restrict the model to using (one of) them during generation. In particular, given a new input $x$, we may generate by computing
\begin{align} \label{eq:templategen}
    \hat{y}^{(i)} = \argmax_{y'} p(y', z^{(i)} \given x),
\end{align}
which gives us a generation $\hat{y}^{(i)}$ for each extracted template $z^{(i)}$. For example, the generation in Figure~\ref{fig:e2eex} is obtained by maximizing \eqref{eq:templategen} with $x$ set to the database in Figure~\ref{fig:e2eex} and $z^{(i)}$ set to the template extracted in Figure~\ref{fig:segmentation}. 
In practice, the $\argmax$ in \eqref{eq:templategen} will be intractable to calculate exactly due to the use of RNNs in defining the emission distribution, and so we approximate it with a constrained beam search. This beam search looks very similar to that typically used with RNN decoders, except the search occurs only over a segment, for a particular latent state $k$. 


\subsection{Discussion}
Returning to the discussion of controllability and interpretability, we note that 
with the proposed model (a) it is possible to explicitly force the generation to use a chosen template $z^{(i)}$, which is itself automatically learned from training data, and (b) that every \textit{segment} in the generated $\hat{y}^{(i)}$ is typed by its corresponding latent variable. We explore these issues 
empirically in Section~\ref{sec:qualitative}.


We also note that these properties may be useful for other text applications, and that they offer an additional perspective on how to approach latent variable modeling for text. Whereas there has been much recent interest in learning continuous latent variable representations for text (see Section~\ref{sec:relatedwork}), it has been somewhat unclear what the latent variables to be learned are intended to capture. On the other hand, the latent, template-like structures we induce here represent a plausible, probabilistic latent variable story, and allow for a more controllable method of generation. 

Finally, we highlight one significant possible issue with this model -- the assumption that segments are independent of each other given the corresponding latent variable and $x$. Here we note that the fact that we are allowed to condition on $x$ is quite powerful. Indeed, a clever encoder could capture much of the necessary interdependence between the segments to be generated (e.g., the correct determiner for an upcoming noun phrase) in its encoding, allowing the segments themselves to be decoded more or less independently, given $x$. 

\section{Data and Methods}

Our experiments apply the approach outlined above to two recent, data-driven NLG tasks. 

\subsection{Datasets}
\label{sec:datasets}
Experiments use the E2E~\citep{novikova2017e2e} and WikiBio~\citep{lebret2016neural} datasets, examples of which are shown in Figures~\ref{fig:e2eex} and \ref{fig:wbex}, respectively. The former dataset, used for the 2018 E2E-Gen Shared Task, contains approximately 50K total examples, and uses 945 distinct word types, and the latter dataset contains approximately 500K examples and uses approximately 400K word types. Because our emission model uses a word-level copy mechanism, any record with a phrase consisting of $n$ words as its value is replaced with $n$ positional records having a single word value, following the preprocessing of \citet{lebret2016neural}. For example, ``type[coffee shop]'' in Figure~\ref{fig:e2eex} becomes ``type-1[coffee]'' and ``type-2[shop].''

For both datasets we compare with published encoder-decoder models, as well as with direct template-style baselines. 
The E2E task is evaluated in terms of BLEU~\citep{papineni02bleu}, NIST~\citep{belz2006comparing}, ROUGE~\citep{lin2004rouge}, CIDEr~\citep{vedantam2015cider}, and METEOR~\citep{banerjee2005meteor}.\footnote{We use the official E2E NLG Challenge scoring scripts at \url{https://github.com/tuetschek/e2e-metrics}.} The benchmark system for the task is an encoder-decoder style system followed by a reranker, proposed by \citet{duvsek2016sequence}. We compare to this baseline, as well as to a simple but competitive non-parametric template-like baseline (``SUB'' in tables), which selects a training sentence with records that maximally overlap (without including extraneous records) the unseen set of records we wish to generate from; ties are broken at random. Then, word-spans in the chosen training sentence are aligned with records by string-match, and replaced with the corresponding fields of the new set of records.\footnote{For categorical records, like ``familyFriendly'', which cannot easily be aligned with a phrase, we simply select only candidate training sentences with the same categorical value.}

The WikiBio dataset is evaluated in terms of BLEU, NIST, and ROUGE, and we compare with the systems and baselines implemented by \citet{lebret2016neural}, which include two neural, encoder-decoder style models, as well as a Kneser-Ney, templated baseline.

\subsection{Model and Training Details}
\label{sec:details}
We first emphasize two additional methodological details important for obtaining good performance.

\vspace{-0.1cm}
\paragraph{Constraining Learning}
We were able to learn more plausible segmentations of $y$ by constraining the model to respect word spans $y_{t+1:t+l}$ that appear in some record $r_j \nicein x$. We accomplish this by giving zero probability (within the backward recurrences in Section~\ref{sec:model}) to any segmentation that splits up a sequence $y_{t+1:t+l}$ that appears in some $r_j$, or that includes $y_{t+1:t+l}$ as a subsequence of another sequence. Thus, we maximize \eqref{eq:marginal} subject to these hard constraints.

\vspace{-0.1cm}
\paragraph{Increasing the Number of Hidden States}
While a larger $K$ allows for a more expressive latent model, computing $K$ emission distributions over the vocabulary can be prohibitively expensive. We therefore tie the emission distribution between multiple 
states, while allowing them to have a different transition distributions.

\vspace{0.15cm}
We give additional architectural details of our model in the Supplemental Material; here we note that we use an MLP to embed $\boldr_j \nicein \reals^d$, and a 1-layer LSTM~\citep{hochreiter1997lstm} in defining our emission distributions. In order to reduce the amount of memory used, we restrict our output vocabulary (and thus the height of the matrix $\boldW$ in Section~\ref{sec:model}) to only contain words in $y$ that are \textit{not} present in $x$; any word in $y$ present in $x$ is assumed to be copied. In the case where a word $y_t$ appears in a record $r_j$ (and could therefore have been copied), the input to the LSTM at time $t+1$ is computed using information from $r_j$; if there are multiple $r_j$ from which $y_t$ could have been copied, the computed representations are simply averaged.

For all experiments, we set $d \niceq 300$ and $L \niceq 4$. At generation time, we select the 100 most common templates $z^{(i)}$, perform beam search with a beam of size 5, and select the generation with the highest overall joint probability. 

For our E2E experiments, our best non-autoregressive model has 55 ``base'' states, duplicated 5 times, for a total of $K \niceq 275$ states, and our best autoregressive model uses $K \niceq 60$ states, without any duplication. For our WikiBio experiments, both our best non-autoregressive and autoregressive models uses 45 base states duplicated 3 times, for a total of $K \niceq 135$ states. In all cases, $K$ was chosen based on BLEU performance on held-out validation data. Code implementing our models is available at~\url{https://github.com/harvardnlp/neural-template-gen}.


\section{Results}

\begin{table}[t!]
\small
\centering
\begin{tabular}{@{}lc@{\hspace{2mm}}c@{\hspace{2mm}}c@{\hspace{2mm}}c@{\hspace{2mm}}c@{}}
\toprule
 & BLEU & NIST & ROUGE & CIDEr & METEOR \\
\midrule
& & & Validation & & \\
\midrule
D\&J & 69.25 & 8.48 & 72.57 & 2.40 & 47.03 \\
SUB  & 43.71 & 6.72 & 55.35 & 1.41 & 37.87 \\
NTemp & 64.53 & 7.66 & 68.60 & 1.82 & 42.46 \\
NTemp+AR & 67.07 & 7.98 & 69.50 & 2.29 & 43.07 \\
\midrule
& & & Test & & \\
\midrule
D\&J & 65.93 & 8.59 & 68.50 & 2.23 & 44.83 \\
SUB & 43.78 & 6.88 & 54.64 & 1.39 & 37.35 \\
NTemp    & 55.17 & 7.14 & 65.70 & 1.70 & 41.91 \\
NTemp+AR & 59.80 & 7.56 & 65.01 & 1.95 & 38.75 \\
\bottomrule
\end{tabular}
\caption{Comparison of the system of \citet{duvsek2016sequence}, which forms the baseline for the E2E challenge, a non-parametric, substitution-based baseline (see text), and our HSMM models (denoted ``NTemp'' and ``NTemp+AR'' for the non-autoregressive and autoregressive versions, resp.) on the validation and test portions of the E2E dataset. ``ROUGE'' is ROUGE-L. Models are evaluated using the official E2E NLG Challenge scoring scripts.}
\label{tab:e2e}
\end{table}

\begin{table}[t!]
\small
\centering
\begin{tabular}{@{}lccc@{}}
\toprule
 & BLEU & NIST & ROUGE-4\\
\midrule
Template KN $\dagger$ & 19.8 & 5.19 & 10.7 \\
NNLM (field) $\dagger$ & 33.4 & 7.52 & 23.9 \\
NNLM (field \& word) $\dagger$ & 34.7 & 7.98 & 25.8 \\
NTemp &  34.2 & 7.94 & 35.9 \\
NTemp+AR &  34.8 & 7.59 & 38.6 \\
\midrule
Seq2seq~\citep{liu2018table} & 43.65 & - & 40.32 \\
\bottomrule
\end{tabular}
\caption{Top: comparison of the two best neural systems of \citet{lebret2016neural}, their templated baseline, and our HSMM models (denoted ``NTemp'' and ``NTemp+AR'' for the non-autoregressive and autoregressive versions, resp.) on the test portion of the WikiBio dataset. Models marked with a {$\dagger$} are from \citet{lebret2016neural}, and following their methodology we use ROUGE-4. Bottom: state-of-the-art seq2seq-style results from \citet{liu2018table}.}
\label{tab:wb}
\end{table}


Our results on automatic metrics are shown in Tables~\ref{tab:e2e} and \ref{tab:wb}. In general, we find that the templated baselines underperform neural models, whereas our proposed model is fairly competitive with neural models, and sometimes even outperforms them. On the E2E data, for example, we see in Table~\ref{tab:e2e} that the SUB baseline, despite having fairly impressive performance for a non-parametric model, fares the worst. The neural HSMM models are largely competitive with the encoder-decoder system on the validation data, despite offering the benefits of interpretability and controllability; however, the gap increases on test. 


Table~\ref{tab:wb}  evaluates our system's performance on the test portion of the WikiBio dataset, comparing with the systems and baselines implemented by \citet{lebret2016neural}.  Again for this dataset we see that their templated Kneser-Ney model underperforms on the automatic metrics, and that neural models improve on these results. Here the HSMMs are competitive with the best model of \citet{lebret2016neural}, and even outperform it on ROUGE. We emphasize, however, that recent, sophisticated approaches to encoder-decoder style database-to-text generation have since surpassed the results of \citet{lebret2016neural} and our own, and we show the recent seq2seq style results of \citet{liu2018table}, who use a somewhat larger model, at the bottom of Table~\ref{tab:wb}.

\subsection{Qualitative Evaluation}
\label{sec:qualitative}
We now qualitatively demonstrate that our generations are controllable and interpretable.


\begin{table}[t]
\small
\centering
\begin{tabular}{@{}l@{}}
\toprule
\textbf{Travellers Rest Beefeater} \\
\midrule 
\begin{tabular}[c]{@{}l@{}}
name[Travellers Rest Beefeater], customerRating[3 out of 5], \\ area[riverside], near[Raja Indian Cuisine]\\
\end{tabular}         \\ 
\midrule
\begin{tabular}[c]{@{}l@{}}
1. [Travellers Rest Beefeater]$_{55}$ [is a]$_{59}$ [3 star]$_{43}$ \\
\; {[restaurant]}$_{11}$ [located near]$_{25}$ [Raja Indian Cuisine]$_{40}$ [.]$_{53}$ \\
2. [Near]$_{31}$ [riverside]$_{29}$ [,]$_{44}$ [Travellers Rest Beefeater]$_{55}$ \\
\; {[serves]}$_{3}$ [3 star]$_{50}$ [food]$_{1}$ [.]$_{2}$ \\
3. [Travellers Rest Beefeater]$_{55}$ [is a]$_{59}$ [restaurant]$_{12}$ \\
\; {[providing]}$_{3}$ [riverside]$_{50}$ [food]$_{1}$ [and has a]$_{17}$ \\
\; {[3 out of 5]}$_{26}$ [customer rating]$_{16}$ [.]$_{2}$ [It is]$_{8}$ [near]$_{25}$ \\
\; {[Raja Indian Cuisine]}$_{40}$ [.]$_{53}$ \\
4. [Travellers Rest Beefeater]$_{55}$ [is a]$_{59}$ [place to eat]$_{12}$ \\ 
\; {[located near]}$_{25}$ {[Raja Indian Cuisine]}$_{40}$ [.]$_{53}$ \\
5. [Travellers Rest Beefeater]$_{55}$ [is a]$_{59}$ {[3 out of 5]}$_{5}$ \\ 
\; {[rated]}$_{32}$ [riverside]$_{43}$ [restaurant]$_{11}$ [near]$_{25}$ \\
\; {[Raja Indian Cuisine]}$_{40}$ [.]$_{53}$
\end{tabular} \\
\bottomrule
\end{tabular}
\caption{Impact of varying the template $z^{(i)}$ for a single $x$ from the E2E validation data; generations are annotated with the segmentations of the chosen  $z^{(i)}$. Results were obtained using the NTemp+AR model from Table~\ref{tab:e2e}.}
\label{tab:e2econtrol}
\end{table}

\begin{table*}[t]
\small
\centering
\begin{tabular}{@{}l@{}}
\toprule
\textbf{kenny warren} \\
\midrule 
\begin{tabular}[c]{@{}l@{}}
\textbf{name:} kenny warren, \textbf{birth date:} 1 april 1946, \textbf{birth name:} kenneth warren deutscher, \textbf{birth place:} brooklyn, new york, \\ \textbf{occupation:} ventriloquist, comedian, author, \textbf{notable work:} book - the revival of ventriloquism in america \\
\end{tabular}         \\ 
\midrule
\begin{tabular}[c]{@{}l@{}}
1. [kenneth warren deutscher]$_{132}$ [ ( ]$_{75}$ [born]$_{89}$ [april 1, 1946]$_{101}$ [ ) ]$_{67}$ [is an american]$_{82}$ [author]$_{20}$ [and]$_{1}$ \\
\; {[ventriloquist and comedian]}$_{69}$ [.]$_{88}$ \\
2. [kenneth warren deutscher]$_{132}$ [ ( ]$_{75}$ [born]$_{89}$ [april 1, 1946]$_{101}$ [ ) ]$_{67}$ [is an american]$_{82}$ [author]$_{20}$ \\
\; {[best known for his]}$_{95}$ [the revival of ventriloquism]$_{96}$ [.]$_{88}$ \\
3. [kenneth warren]$_{16}$ [``kenny'' warren]$_{117}$ [ ( ]$_{75}$ [born]$_{89}$ [april 1, 1946]$_{101}$ [ ) ]$_{67}$ [is an american]$_{127}$ \\
\; {[ventriloquist, comedian]}$_{28}$ [.]$_{133}$ \\
4. [kenneth warren]$_{16}$ [``kenny'' warren]$_{117}$ [ ( ]$_{75}$ [born]$_{89}$ [april 1, 1946]$_{101}$ [ ) ]$_{67}$ [is a]$_{104}$ [new york]$_{98}$ [author]$_{20}$ [.]$_{133}$ \\
5. [kenneth warren deutscher]$_{42}$ [is an american]$_{82}$ [ventriloquist, comedian]$_{118}$ [based in]$_{15}$ [brooklyn, new york]$_{84}$ [.]$_{88}$
\end{tabular} \\
\bottomrule
\end{tabular}
\caption{Impact of varying the template $z^{(i)}$ for a single $x$ from the WikiBio validation data; generations are annotated with the segmentations of the chosen  $z^{(i)}$. Results were obtained using the NTemp model from Table~\ref{tab:wb}.}
\label{tab:wbcontrol}
\end{table*}


\paragraph{Controllable Diversity}
One of the powerful aspects of the proposed approach to generation is that we can manipulate the template $z^{(i)}$ while leaving the database $x$ constant, which allows for easily controlling aspects of the generation. In Table~\ref{tab:e2econtrol}  we show the generations produced by our model for five different neural template sequences $z^{(i)}$, while fixing $x$. There, the segments in each generation are annotated with the latent states determined by the corresponding $z^{(i)}$. We see that these templates can be used to affect the word-ordering, as well as which fields are mentioned in the generated text. Moreover, because the discrete states align with particular fields (see below), it is generally simple to automatically infer to which fields particular latent states correspond, allowing users to choose which template best meets their requirements. We emphasize that this level of controllability is much harder to obtain for encoder-decoder models, since, at best, a large amount of sampling would be required to avoid generating around a particular mode in the conditional distribution, and even then it would be difficult to control the sort of generations obtained. 

\paragraph{Interpretable States}
Discrete states also provide a method for interpreting the generations produced by the system, since each segment is explicitly typed by the current 
hidden state of the model. Table~\ref{tab:wbcontrol} shows the impact of varying the template $z^{(i)}$ for a single $x$ from the WikiBio dataset. While there is in general surprisingly little stylistic variation in the WikiBio data itself, there is variation in the information discussed, and the templates capture this. Moreover, we see that particular discrete states correspond in a consistent way to particular pieces of information, allowing us to align states with particular field types. For instance, birth names have the same hidden state (132), as do names (117), nationalities (82), birth dates (101), and occupations (20). 

To demonstrate empirically that the learned states indeed align with field types, we calculate the average purity of the discrete states learned for both datasets in Table~\ref{tab:alignment}. In particular, for each discrete state for which the majority of its generated words appear in some $r_j$, the \textit{purity} of a state's record type alignment is calculated as the percentage of the state's words that come from the most frequent record type the state represents. This calculation was carried out over training examples that belonged to one of the top 100 most frequent templates. Table~\ref{tab:alignment} indicates that discrete states learned on the E2E data are quite pure. Discrete states learned on the WikiBio data are less pure, though still rather impressive given that there are approximately 1700 record types represented in the WikiBio data, and we limit the number of states to 135. Unsurprisingly, adding autoregressiveness to the model decreases purity on both datasets, since the model may rely on the autoregressive RNN for typing, in addition to the state's identity.


\begin{table}[t!]
\small
\centering
\begin{tabular}{@{}lcc}
\toprule
 & NTemp & NTemp+AR\\
\midrule
E2E & 81.7 (17.9) & 81.2 (15.7)  \\
WikiBio & 37.5 (18.9) & 36.3 (20.2) \\
\bottomrule
\end{tabular}
\caption{Empirical analysis of the average purity of discrete states learned on the E2E and WikiBio datasets, for the NTemp and NTemp+AR models. Average purities are given as percents, and standard deviations follow in parentheses. See the text for full description of this calculation.}
\label{tab:alignment}
\end{table}


\section{Conclusion and Future Work}

We have developed a neural, template-like generation model based on an HSMM decoder, which can be learned tractably by backpropagating through a dynamic program. The method allows us to extract template-like latent objects in a principled way in the form of state sequences, and then generate with them. This approach scales to large-scale text datasets and is nearly competitive with encoder-decoder models. More importantly, this approach allows for controlling the diversity of generation and for producing interpretable states during generation.  We view this work both as the first step towards learning discrete latent variable template models for more difficult generation tasks, as well as a different perspective on learning latent variable text models in general.  Future work will examine encouraging the model to learn maximally different (or minimal) templates, which our objective does not explicitly encourage, templates of larger textual phenomena, such as paragraphs and documents, and hierarchical templates.

\section*{Acknowledgments}
SW gratefully acknowledges the support of a Siebel Scholars award. AMR gratefully acknowledges the support of NSF CCF-1704834, Intel Research, and Amazon AWS Research grants.

\nocite{tran2016unsupervised}

\bibliography{gen2}
\bibliographystyle{acl_natbib_nourl}

\appendix
\section{Supplemental Material}
\label{sec:supplemental}
\subsection{Additional Model and Training Details}
\paragraph{Computing $\boldr_j$}
A record $r_j$ is represented by embedding a feature for its type, its position, and its word value in $\reals^d$, and applying an MLP with ReLU nonlinearity~\citep{nair2010rectified} to form $\boldr_j \nicein \reals^d$, similar to \citet{yang2016reference} and \citet{wiseman2017challenges}.

\paragraph{LSTM Details}
The initial cell and hidden-state values for the decoder LSTM are given by $\boldQ_1 \boldx_a$ and $\tanh(\boldQ_2 \boldx_a)$, respectively, where $\boldQ_1, \boldQ_2 \nicein \reals^{d \times d}$. 

When a word $y_t$ appears in a record $r_j$, the input to the LSTM at time $t+1$ is computed using an MLP with ReLU nonlinearity over the concatenation of the embeddings for $r_j$'s record type, word value, position, and a feature for whether it is the final position for the type. If there are multiple $r_j$ from which $y_t$ could have been copied, the computed representations are averaged. At test time, we use the MAP $r_j$ to compute the input, even if there are multiple matches. For $y_t$ which could not have been copied, the input to the LSTM at time $t+1$ is computed using the same MLP over $y_t$ and three dummy features.

For the autoregressive HSMM, an additional 1-layer LSTM with $d$ hidden units is used. We experimented with having the autoregressive HSMM consume either tokens $y_{1:t}$ in predicting $y_{t+1}$, or the average embedding of the field \textit{types} corresponding to copied tokens in $y_{1:t}$. The former worked slightly better for the WikiBio dataset (where field types are more ambiguous), while the latter worked slightly better for the E2E dataset.

\paragraph{Transition Distribution}
The function $\boldC(\boldx_u)$, which produces hidden state embeddings conditional on the source, is defined as $\boldC(\boldx_u) \niceq \boldU_2 (\mathrm{ReLU} (\boldU_1 \boldx_u))$, where $\boldU_1 \nicein \reals^{m_3 \times d}$ and $\boldU_2 \nicein \reals^{K \times m_2 \times m_3}$; $\boldD(x)$ is defined analogously. For all experiments, $m_1 \niceq 64$, $m_2 \niceq 32$, and $m_3 \niceq 64$.

\paragraph{Optimization}
We train with SGD, using a learning rate of 0.5 and decaying by 0.5 each epoch after the first epoch in which validation log-likelihood fails to increase. When using an autoregressive HSMM, the additional LSTM is optimized only after the learning rate has been decayed. We regularize with Dropout~\citep{srivastava2014dropout}.

\subsection{Additional Learned Templates}
In Tables~\ref{tab:moare2etemps} and \ref{tab:moarwbtemps} we show visualizations of additional templates learned on the E2E and WikiBio data, respectively, by both the non-autoregressive and autoregressive HSMM models presented in the paper. For each model, we select a set of five dissimilar templates in an iterative way by greedily selecting the next template (out of the 200 most frequent) that has the highest percentage of states that do not appear in the previously selected templates; ties are broken randomly. Individual states within a template are visualized using the three most common segments they generate.

\newcommand{\tslot}[3]{$\textcolor{gray}{|} \ \substack{\text{#1}\\\text{#2}\\\text{#3}\\\dots}$}

\begin{table*}
\centering
\begin{tabular}{ll}
\toprule
1. & \tslot{The Waterman}{The Golden Palace}{Browns Cambridge} \tslot{is a}{is an}{is a family friendly} \tslot{Italian}{French}{fast food} \tslot{restaurant}{pub}{place} \tslot{with a}{with}{with an} \tslot{average}{high}{low} \tslot{customer rating}{price range}{rating}$\textcolor{gray}{|}$. \\
\addlinespace[2mm]
2. & \tslot{There is a}{There is a cheap}{There is an} \tslot{restaurant}{coffee shop}{French restaurant} \tslot{The Mill}{Bibimbap House}{The Twenty Two} \tslot{located in the}{located on the}{located north of the} \tslot{centre of the city}{river}{city centre} \tslot{that serves}{serving}{that provides} \\
\addlinespace[1mm]
& \qquad \tslot{fast food}{sushi}{take-away deliveries}$\textcolor{gray}{|}$. \\
\addlinespace[2mm]
3. & \tslot{The Olive Grove}{The Punter}{The Cambridge Blue} \tslot{}{restaurant}{pub} \tslot{serves}{offers}{has} \tslot{fast food}{sushi}{take-away deliveries}$\textcolor{gray}{|}$. \\
\addlinespace[2mm]
4. & \tslot{The}{Child friendly}{The average priced} \tslot{restaurant}{coffee shop}{French restaurant} \tslot{The Mill}{Bibimbap House}{The Twenty Two} \tslot{serves}{offers}{has} \tslot{English}{Indian}{Italian} \tslot{food}{cuisine}{dishes}$\textcolor{gray}{|}$. \\
\addlinespace[2mm]
5. & \tslot{The Strada}{The Dumpling Tree}{Alimentum} \tslot{provides}{serves}{offers} \tslot{Indian}{Chinese}{English} \tslot{food in the}{food at a}{food and has a} \tslot{customer rating of}{price range of}{rating of} \tslot{1 out of 5}{average}{5 out of 5}$\textcolor{gray}{|}$. \\
\midrule
1. & \tslot{The Eagle}{The Golden Curry}{Zizzi} \tslot{provides}{providing}{serves} \tslot{Indian}{Chinese}{English} \tslot{food}{cuisine}{Food} \tslot{in the}{with a}{and has a} \tslot{high}{moderate}{average} \tslot{price range}{customer rating}{rating}$\textcolor{gray}{|}$. \tslot{It is}{They are}{It's} \tslot{near}{located in the}{located near} \\
\addlinespace[1mm]
& \qquad \tslot{riverside}{city centre}{Cafe Sicilia} $\textcolor{gray}{|}$. \tslot{Its customer rating is}{It has a}{The price range is} \tslot{1 out of 5}{average}{high}$\textcolor{gray}{|}$. \\
\addlinespace[2mm]
2. & \tslot{Located near}{Located in the}{Near} \tslot{The Portland Arms}{riverside}{city centre} \tslot{is an}{is a family friendly}{there is a} \tslot{Italian}{fast food}{French} \tslot{restaurant called}{place called}{restaurant named} \tslot{The Waterman}{Cocum}{Loch Fyne}$\textcolor{gray}{|}$. \\
\addlinespace[2mm]
3. & \tslot{A}{An}{A family friendly} \tslot{Italian}{fast food}{French} \tslot{restaurant}{pub}{coffee shop} \tslot{is}{called}{named} \tslot{The Waterman}{Cocum}{Loch Fyne}$\textcolor{gray}{|}$. \\
\addlinespace[2mm]
4. & \tslot{Located near}{Located in the}{Near} \tslot{The Portland Arms}{riverside}{city centre}$\textcolor{gray}{|}$ , \tslot{The Eagle}{The Golden Curry}{Zizzi} \tslot{is a}{is a family friendly}{is an} \tslot{cheap}{family-friendly}{family friendly} \tslot{Italian}{fast food}{French} \tslot{restaurant}{pub}{coffee shop}$\textcolor{gray}{|}$. \\
\addlinespace[2mm]
5. & \tslot{A}{An}{A family friendly} \tslot{Italian}{fast food}{French} \tslot{restaurant}{pub}{coffee shop} \tslot{near}{located in the}{located near} \tslot{riverside}{city centre}{Cafe Sicilia} \tslot{is}{called}{named} \tslot{The Waterman}{Cocum}{Loch Fyne}$\textcolor{gray}{|}$. \\
\bottomrule
\end{tabular}
\caption{Five templates extracted from the E2E data with the NTemp model (top) and the Ntemp+AR model (bottom).}
\label{tab:moare2etemps}
\end{table*}

\begin{table*}
\centering
\begin{tabular}{ll}
\toprule
1. & \tslot{william henry}{george augustus frederick}{marie anne de bourbon} \tslot{(}{was (}{;} \tslot{born}{born on}{born 1} \tslot{1968}{1960}{1970} \tslot{)}{])}{]} \tslot{is an american}{is a russian}{was an american} \tslot{politician}{actor}{football player}$\textcolor{gray}{|}$. \\
\addlinespace[2mm]
2. & \tslot{sir}{captain}{lieutenant} \tslot{john herbert}{hartley}{donald charles cameron} \tslot{was a}{was a british}{was an english} \tslot{world war i}{world war}{first world war} \tslot{national team}{organization}{super league}$\textcolor{gray}{|}$. \\
\addlinespace[2mm]
3. & \tslot{john herbert}{hartley}{donald charles cameron} \tslot{is a}{was a}{is an} \tslot{indie rock}{death metal}{ska} \tslot{band}{midfielder}{defenceman} \tslot{from}{for}{based in} \tslot{australia}{los angeles, california}{chicago}$\textcolor{gray}{|}$. \\
\addlinespace[2mm]
4. & \tslot{john herbert}{hartley}{donald charles cameron} \tslot{was a}{is a}{is a former} \tslot{american}{major league baseball}{australian} \tslot{football}{professional baseball}{professional ice hockey} \tslot{midfielder}{defender}{goalkeeper}$\textcolor{gray}{|}$. \\
\addlinespace[2mm]
5. & \tslot{james}{william john}{william} \tslot{`` billy '' wilson}{smith}{`` jack '' henry} $\textcolor{gray}{|}$ ( \tslot{1900}{c. 1894}{1913} $\textcolor{gray}{|}$ -- \tslot{france}{budapest}{buenos aires} $\textcolor{gray}{|}$ ) \tslot{is an american}{is an english}{was an american} \tslot{footballer}{professional footballer}{rules footballer} \\
\addlinespace[1mm]
& \qquad  \tslot{who plays for}{who currently plays for}{who played with} \tslot{paganese}{south melbourne}{fc dynamo kyiv} \tslot{in the}{of the}{and the} \tslot{vicotiral football league}{national football league}{australian football league} $\textcolor{gray}{|}$ ( \tslot{vfl}{nfl}{afl} $\textcolor{gray}{|}$ ) $\textcolor{gray}{|}$. \\
\midrule
1. & \tslot{aftab ahmed}{anderson da silva}{david jones} \tslot{}{(}{;} \tslot{born}{born on}{born 1} \tslot{1951}{1970}{1974} \tslot{}{)}{]} \tslot{is an american}{was an american}{is an english} \tslot{actor}{actress}{cricketer}$\textcolor{gray}{|}$. \\
\addlinespace[2mm]
2. & \tslot{aftab ahmed}{anderson da silva}{david jones} \tslot{was a}{is a former}{is a} \tslot{world war i}{liberal}{baseball} \tslot{member of the}{party member of the}{recipient of the} \tslot{austrian}{pennsylvania}{montana} \tslot{house of representatives}{legislature}{senate}$\textcolor{gray}{|}$. \\
\addlinespace[2mm]
3. & \tslot{adjutant}{lieutenant}{captain} \tslot{aftab ahmed}{anderson da silva}{david jones} \tslot{was a}{is a former}{is a} \tslot{world war i}{liberal}{baseball} \tslot{member of the}{party member of the}{recipient of the} \tslot{knesset}{scottish parliament}{fc lokomotiv liski}$\textcolor{gray}{|}$. \\
\addlinespace[2mm]
4. & \tslot{william}{john william}{james ``} \tslot{`` billy '' watson}{smith}{jim '' edward} $\textcolor{gray}{|}$ ( \tslot{1913}{c. 1900}{1913} \tslot{--}{in}{-} \tslot{1917}{surrey, england}{british columbia} $\textcolor{gray}{|}$ ) \tslot{was an american}{was an australian}{is an american} \tslot{football player}{rules footballer}{defenceman}  \\
\addlinespace[1mm]
& \qquad \tslot{who plays for}{who currently plays for}{who played with} \tslot{collingwood}{st kilda}{carlton} \tslot{in the}{of the}{and the} \tslot{victorial football league}{national football league}{australian football league} $\textcolor{gray}{|}$ ( \tslot{vfl}{afl}{nfl} $\textcolor{gray}{|}$ ) $\textcolor{gray}{|}$. \\
\addlinespace[2mm]
5. & \tslot{aftab ahmed}{anderson da silva}{david jones} \tslot{is a}{is a former}{is a female} \tslot{member of the}{party member of the}{recipient of the} \tslot{knesset}{scottish parliament}{fc lokomotiv liski} $\textcolor{gray}{|}$. \\
\bottomrule
\end{tabular}
\caption{Five templates extracted from the WikiBio data with the NTemp model (top) and the Ntemp+AR model (bottom).}
\label{tab:moarwbtemps}
\end{table*}

\end{document}